\documentclass[10pt, a4paper]{article}

\usepackage[]{lrec-coling2024} %
\usepackage{enumerate}
\usepackage{microtype}
\usepackage{booktabs}
\usepackage{adjustbox}
\usepackage{multirow}
\usepackage{multicol}
\usepackage{subcaption}
\usepackage{amsfonts}
\usepackage{amsmath}
\usepackage{cleveref}

\title{Can Machine Translation Bridge Multilingual Pretraining and Cross-lingual Transfer Learning?}

\name{Shaoxiong Ji~\textsuperscript{1} \quad Timothee Mickus~\textsuperscript{1} \quad Vincent Segonne~\textsuperscript{2} \quad \textbf{Jörg Tiedemann}~\textsuperscript{1}}
\address{
\textsuperscript{1} University of Helsinki \quad \textsuperscript{2} Universite Grenoble Alpes
\\
\texttt{firstname.lastname@helsinki.fi}
}

\abstract{
 Multilingual pretraining and fine-tuning have remarkably succeeded in various natural language processing tasks. 
 Transferring representations from one language to another is especially crucial for cross-lingual learning. One can expect machine translation objectives to be well suited to fostering such capabilities, as they involve the explicit alignment of semantically equivalent sentences from different languages. 
 This paper investigates the potential benefits of employing machine translation as a continued training objective to enhance language representation learning, bridging multilingual pretraining and cross-lingual applications. 
 We study this question through two lenses: a quantitative evaluation of the performance of existing models and an analysis of their latent representations. 
 Our results show that, contrary to expectations, machine translation as the continued training fails to enhance cross-lingual representation learning in multiple cross-lingual natural language understanding tasks. 
 We conclude that explicit sentence-level alignment in the cross-lingual scenario is detrimental to cross-lingual transfer pretraining, which has important implications for future cross-lingual transfer studies.
 We furthermore provide evidence through similarity measures and investigation of parameters that this lack of positive influence is due to output separability---which we argue is of use for machine translation but detrimental elsewhere.
  \\ \newline \Keywords{Cross-lingual Transfer Learning, Representation Similarity and Explainability, Machine Translation, Multilingual Language Models} }

\begin{document}

\maketitleabstract

\section{Introduction}
\label{sec:introduction}

The successes of pretrained multilingual language models (LM) on cross-lingual tasks have been underscored time and time again \citep[e.g.,]{wu2019beto}, 
and appear all the more surprising that they are often pretrained on datasets comprising multiple languages, without explicit cross-lingual supervision (cf., for instance, \citealp{liu-etal-2020-multilingual-denoising}, though explicit supervision also exists, \citealp{xue-etal-2021-mt5}).
Explicit alignments such as linear mapping~\citep{wang2019cross} and L2 alignment~\citep{cao2020multilingual} between source and target languages do not necessarily improve the quality of cross-lingual representations \citep{wu2020explicit}.

This is somewhat at odds with expectations from earlier studies in machine translation (MT).
In particular, MT systems have historically connected with the concept of an interlingua---a language-independent representation space that MT systems can leverage to perform translation \citep{masterman-1961-semantic,lu-etal-2018-neural}.
As such, MT models are expected to pick up on language-independent semantic features \citep{Tiedemann2018EmergingLS}---though in practice, this shared representation space can be in a trade-off relationship with performance, which benefits from a greater separability of source language representations \citep[e.g.]{chen2023offtarget}. 

\paragraph{Research questions}
This paper investigates whether machine translation as a learning objective can improve performances on zero-shot cross-lingual transfer downstream tasks. 
We expect that MT objectives, as they provide explicit cross-lingual alignments, should benefit cross-lingual transfer tasks.
This paper, therefore, focuses on comparing the cross-lingual abilities of publicly available pretrained models---both MT models trained from scratch and multilingual LMs where pretraining has been continued with an MT objective.
We attempt to establish whether MT training objectives implicitly foster cross-lingual alignment:
\begin{enumerate}[(i)]
    \item Do models (re)trained with the MT objective develop cross-lingual representations? 
    \item Do they generalize well on cross-lingual tasks? 
    \item Which factors impact their performances?
\end{enumerate}

\paragraph{Findings} We find that MT (continued) training objectives do not favor the emergence of cross-lingual alignments more than LM objectives, based on the study on existing publicly available pretrained models. 
We provide evidence from similarity analyses and parameter-level investigations that this is due to separability, which is beneficial in MT but detrimental elsewhere.
We conclude that MT encourages behavior that is not necessarily compatible with high performances in cross-lingual transfer learning.

\section{Experimental protocol}

Our goal is to compare LM and MT models on cross-lingual benchmarks.
We first describe multilingual LMs and MT systems, cross-lingual tasks, and datasets used in our experiments.

\subsection{Publicly available pretrained models}
\label{sec:models}

\paragraph{Multilingual language models}
We study three different multilingual LMs.
The main model we focus on is the multilingual sequence-to-sequence mBART-large model \citep{tang2020multilingual}.
It is pretrained with a denoising objective and covers 50 languages.
It has a 12-layer encoder, a 12-layer decoder, a hidden dimension of 1024, and 16 attention heads, for a total of about 680M parameters.
We also compare the results with masked language models as references by controlling the same level of the number of parameters, mainly the number of transformer layers.
We consider mBERT~\cite{devlin2019bert} and XLM-R~\cite{conneau2020unsupervised} 12-layer base architectures to give a relatively fair comparison.
Nevertheless, for the large mBART architectures, although we only utilize the 12-layer encoder, mBART encoders have roughly 10\% parameters more than mBERT~\cite{devlin2019bert}.

\paragraph{Machine translation model}
We focus on the ``No Language Left Behind'' translation system \citep[`NLLB',][]{costa2022no}.
This model distinguishes itself by using Mixture-of-Experts feedforward sub-layers, intended to ensure that the model can handle inputs from diverse languages.
We use the distilled model with 600M parameters to keep parameter counts roughly consistent with the aforementioned multilingual LMs.

\paragraph{MT as continued pretraining}
Our starting hypothesis is that the MT objective provides an explicit cross-lingual sentence alignment that is likely beneficial for cross-lingual transfer.
A natural, testable consequence of this hypothesis is that further training multilingual LMs with an MT objective should bolster the models' performance on cross-lingual transfer learning benchmarks.
We refer to this sequential training on an MT objective as \emph{continued pretraining} or CP, to distinguish it from task-specific fine-tuning processes.
We use three publicly available mBART models where pretraining was continued on machine translation objectives \citep{tang2020multilingual}: a many-to-many (m2m) which translates between any pair of languages from a pool of 50; a many-to-one (m2o) from any of 49 languages to English; and a one-to-many (o2m) from English to any of 49 languages. 
The continual training of mBART covers a larger number of languages than the down-stream evaluation.
Fine-tuning on a larger set of languages might provide the model with a more diverse linguistic representation. 
We are interested in the hypothesis that this diversity could potentially enhance the model's ability to generalize across languages, even if some of them are not directly involved in the downstream tasks.
Catastrophic forgetting is a significant challenge in continual learning scenarios. However, we stress that this falls beyond the scope of our paper, as our primary focus is on the model training with different learning objectives, and their empirical results on cross-lingual tasks with a further step of training on downstream datasets.

\subsection{Cross-lingual tasks and datasets}
\label{sec:tasks_datasets}

We study the models' performances detailed in \Cref{sec:models} on standard cross-lingual NLP tasks. 
In all cases,  models are trained for the downstream application in one language (usually English), and the trained model is then evaluated in languages other than the language used for training.
We use the XGLUE cross-lingual evaluation benchmark \citep{liang2020XGLUE} and conduct our evaluation on natural language understanding tasks. 
The specific tasks consist of Named Entity Resolution \citep[NER,][]{sang2002IntroductionTT,sang2003IntroductionTT}, 
Part-of-Speech tagging \citep[POS,][]{universal-dependencies}, 
News Classification (NC), natural language inference \citep[XNLI,][]{conneau2018XNLIEC}, 
paraphrase detection \citep[PAWS-X,][]{yang2019PAWSXAC}, 
Query-Ad Matching (QADSM), 
Web Page Ranking (WPR), and 
QA Matching (QAM).
\Cref{tab:statistics} in \Cref{sec:detailed_benchmarks} summarizes the benchmarks used in our study.
For named entity recognition (NER) and web page ranking (WPR), we use the F1 score and normalized discounted cumulative gain (nDCG) as the evaluation metric.
The other tasks use accuracy as the metric.

\subsection{Hyperparameters}
\label{sec:hyperparameters}

We control most experimental settings to enable fair cross-lingual evaluation as much as possible. 
We use 12-layer encoders for each backbone network. 
For optimization, we use the AdamW optimizer~\citep{loshchilov2019decoupled} and learning rate schedule with linear warmup and decay. 
We set the learning rate to $2\times 10^{-5}$ for POS tagging and $5\times 10^{-6}$ for the other tasks.
The max sequence length is 256, and we fine-tune each model for 10 epochs.

\section{Results and analyses}
\label{sec:MT_cross_lingual}

\subsection{Quantitative performance}
\label{sec:perfs}
We first compare the overall performance of the models listed in \Cref{sec:models} on the downstream cross-lingual benchmarks outlined in \Cref{sec:tasks_datasets}.
\Cref{tab:mt-continued-cross-lingual} shows the overall performance by averaging the scores of each language.
XLM-R displays the highest performances on 6 out of 8 tasks, and mBART obtains the best average score on the last two. 
Models continually pretrained on MT (i.e., mBART m2o, mBART o2m, and mBART m2m) perform worse than language models (i.e., mBART) in most cases. 
Multilingual MT models that encode multiple source languages (i.e., m2m and m2o) display comparable or slightly improved performances; for example, mBART m2m outperforms mBERT on PAWS-X and mBART m2o outperforms mBERT on XNLI and QADSM.\footnote{
    The detailed per-language performance is available in \Cref{sec:per-language-results}, \Cref{tab:mt-continued-cross-lingual-per-lang}. 
    In short, mBART and corresponding MT models perform poorly on languages that are unavailable in its training data. 
}
MT models based on mBART achieve satisfactory performance on the English test set in most cases but fail to bridge pretraining and cross-lingual transfer learning in other languages. 
Overall, we find that machine translation as continued pretraining does not improve cross-lingual performance.

\begin{table}[!htbp]
\centering
\scriptsize
\setlength{\tabcolsep}{1.2pt}
\begin{tabular}{llcccccccc}
\toprule
\multicolumn{2}{c}{\multirow{2}{*}{\textbf{Model}}}	& \multicolumn{8}{c}{\textbf{Tasks}} \\
\multicolumn{2}{c}{\multirow{2}{*}{}}&	\textbf{NC}	&	\textbf{XNLI}	&	\textbf{PAWS-X}	&	\textbf{QAM}	&	\textbf{QADSM}	&	\textbf{WPR}	&	\textbf{NER}	&	\textbf{POS}	\\
\midrule
\multirow{3}{*}{\textbf{LM}} 
& mBERT	&	81.3	&	65.2	&	86.6	&	64.6	&	63.1	&	74.4	&	77.5	&	76.0	\\
& XLM-R	&	\textbf{82.1}	&	\textbf{73.5}	&	88.9	&	67.4	&	\textbf{66.9}	&	\textbf{75.3}	&	\textbf{78.7}	&	\textbf{79.7}	\\
& mBART	&	82.1	&	67.6	&	\textbf{89.2}	&	\textbf{67.8}	&	65.5	&	74.7	&	77.7	&	72.7	\\
\midrule
\textbf{MT} & 
NLLB 600M	&	76.0	&	68.3	&	73.4	&	61.5	&	63.9	&	73.7	&	54.2	&	71.4	\\
\midrule
\multirow{3}{*}{\textbf{CP}} 
& mBART m2o	&	80.4	&	65.9	&	85.6	&	63.9	&	63.9	&	73.7	&	61.5	&	70.8	\\
& mBART o2m	&	65.4	&	48.1	&	81.7	&	58.4	&	62.7	&	73.2	&	55.1	&	55.7	\\
& mBART m2m	&	78.3	&	60.2	&	87.2	&	63.2	&	62.8	&	73.7	&	71.9	&	69.7	\\
\bottomrule
\end{tabular}
\caption{Average performance on cross-lingual tasks. We use the base architecture for mBERT and XLM-R. mBART scores are derived from the 12-layer encoder. } %
\label{tab:mt-continued-cross-lingual}
\end{table}

\subsection{Representation similarity}
\label{sec:RSA}
We have established that CP and MT models fare worse than available multilingual LMs, disproving our starting hypothesis.
We now turn to whether these quantitative differences translate into qualitative differences that we can observe in the representational space.
\begin{figure}[!ht]
     \centering
     \begin{subfigure}[b]{0.24\textwidth}
         \centering
         \includegraphics[width=\textwidth]{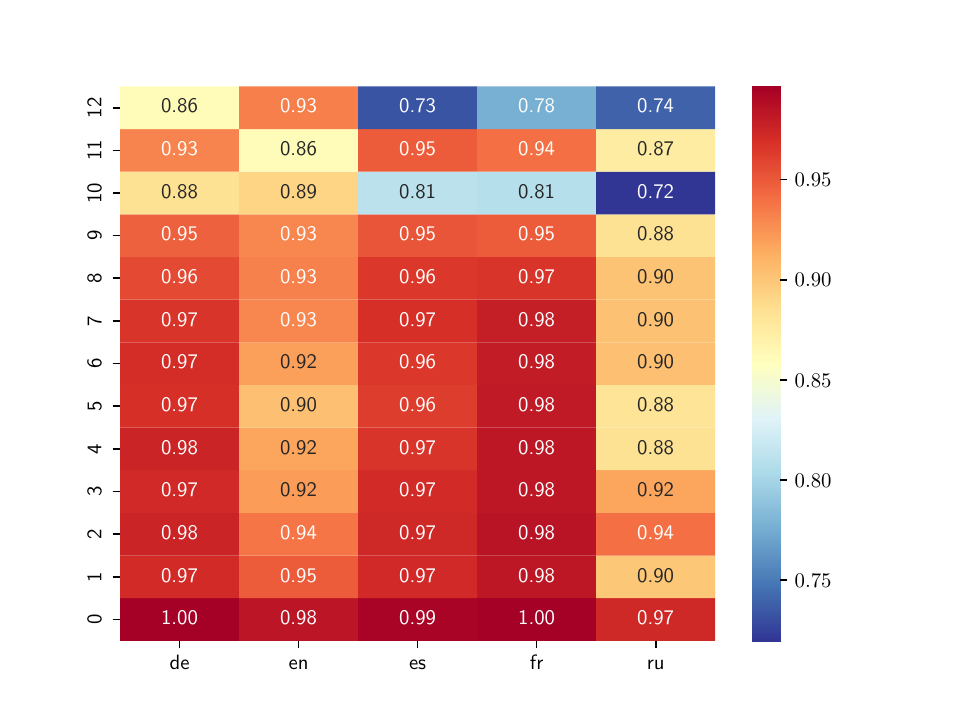}
         \caption{mBART vs mBART m2m}
         \label{fig:mbart-vs-m2m}
     \end{subfigure}
     \hfill
     \begin{subfigure}[b]{0.23\textwidth}
         \centering
         \includegraphics[width=\textwidth]{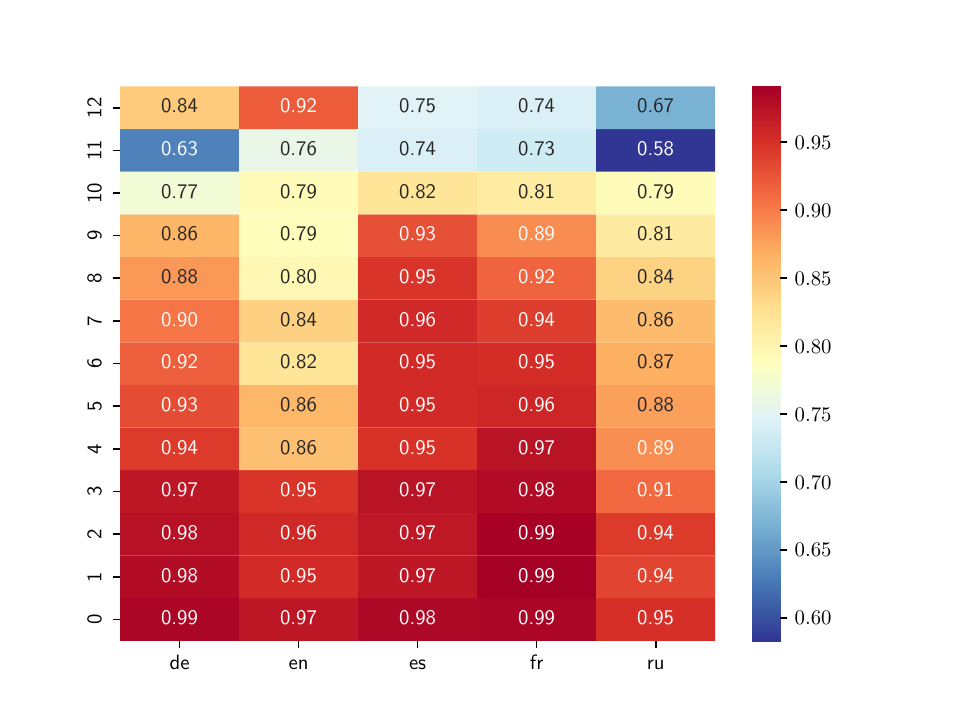}
         \caption{XLM-R vs mBART m2m}
         \label{fig:xlmr-vs-m2m}
     \end{subfigure}
     \hfill
     \begin{subfigure}[b]{0.23\textwidth}
         \centering
         \includegraphics[width=\textwidth]{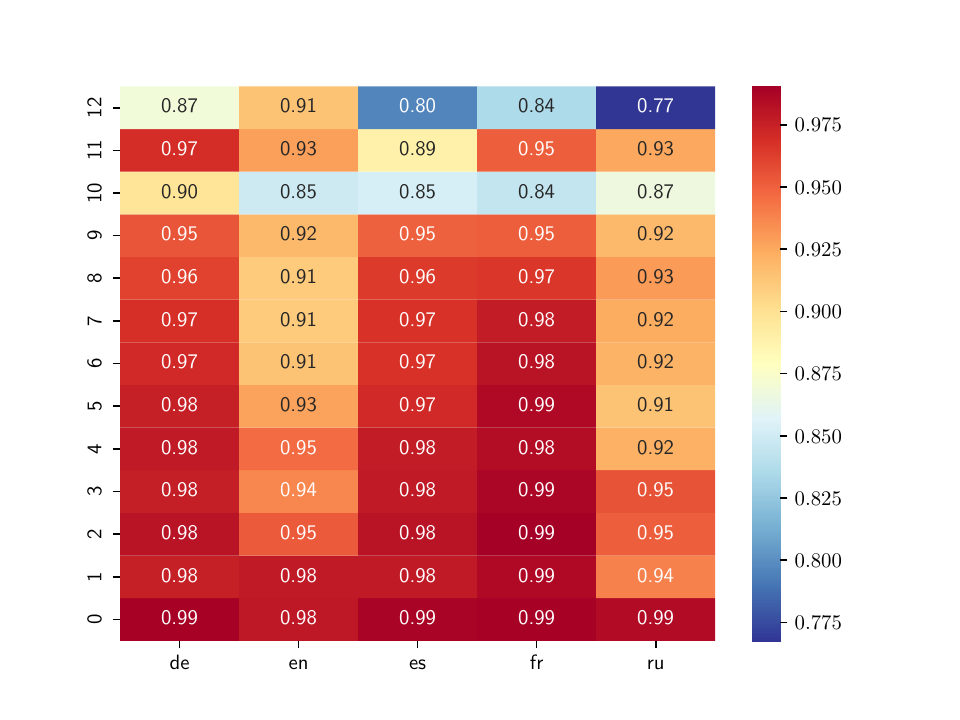}
         \caption{mBART vs mBART m2o}
         \label{fig:mbart-vs-m2o}
     \end{subfigure}
     \hfill
     \begin{subfigure}[b]{0.23\textwidth}
         \centering
         \includegraphics[width=\textwidth]{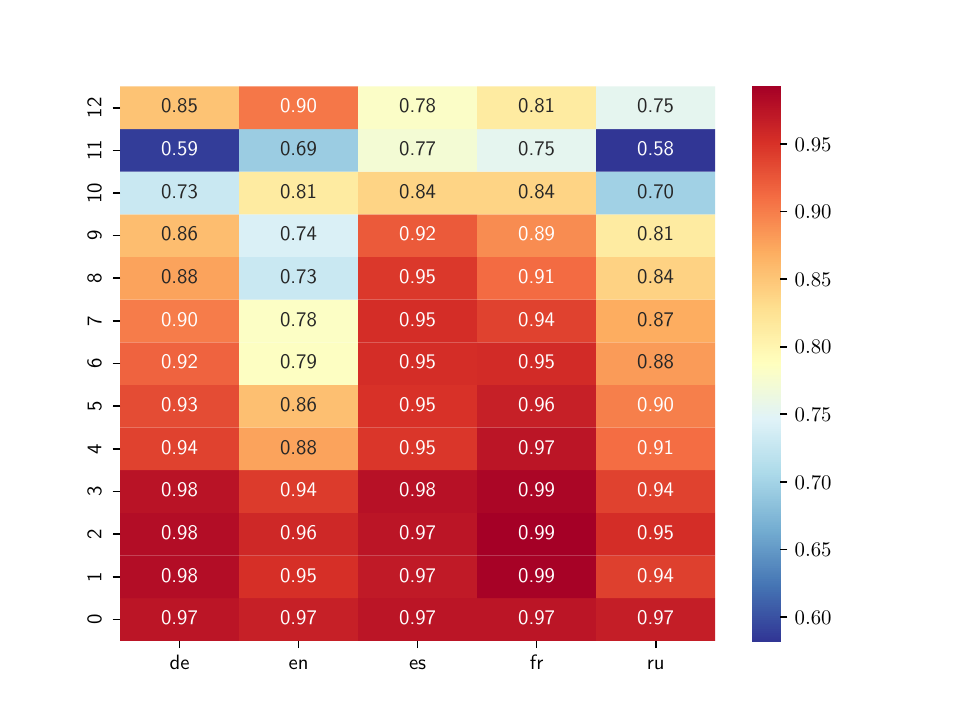}
         \caption{XLM-R vs mBART m2o}
         \label{fig:xlmr-vs-m2o}
     \end{subfigure}
     \hfill
     \begin{subfigure}[b]{0.23\textwidth}
         \centering
         \includegraphics[width=\textwidth]{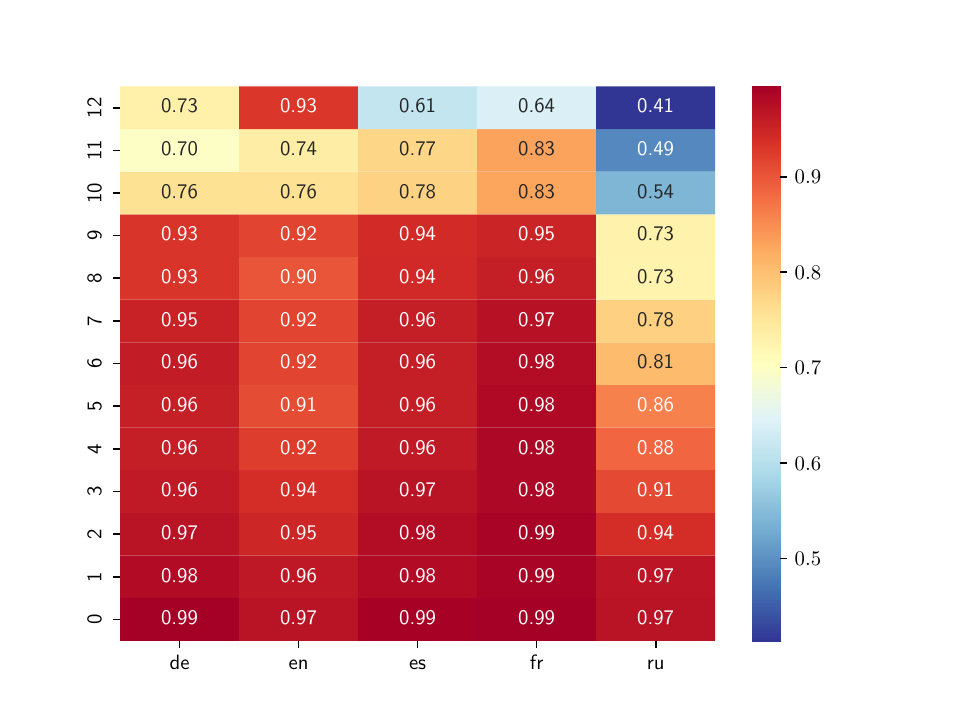}
         \caption{mBART vs mBART o2m}
         \label{fig:mbart-vs-o2m}
     \end{subfigure}
    \hfill
     \begin{subfigure}[b]{0.23\textwidth}
         \centering
         \includegraphics[width=\textwidth]{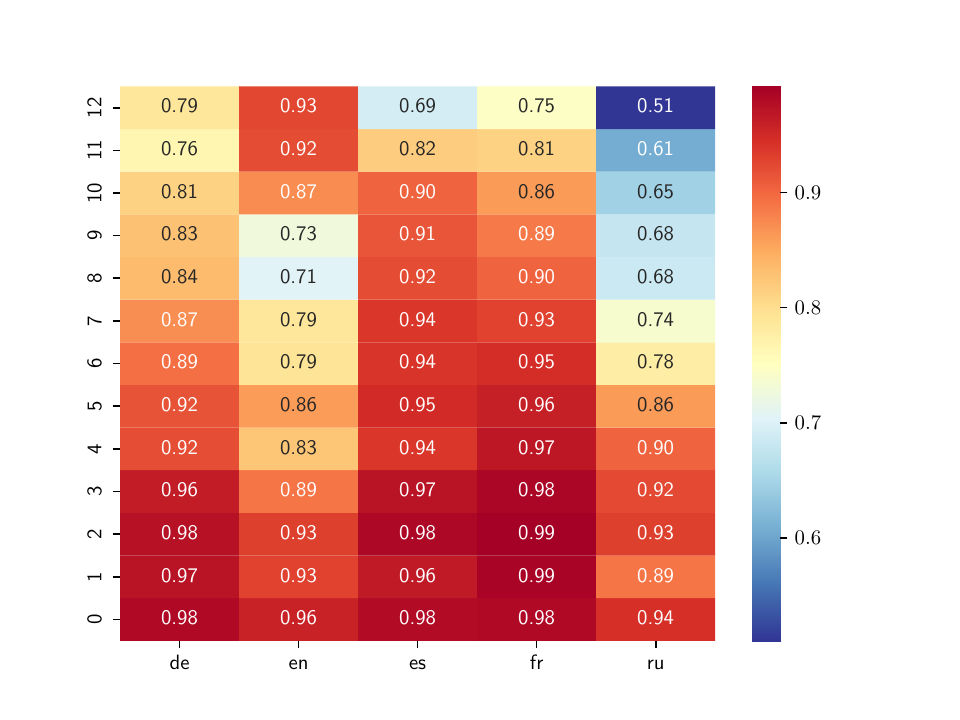}
         \caption{XLM-R vs mBART o2m}
         \label{fig:xlmr-vs-o2m}
     \end{subfigure}
     \begin{subfigure}[b]{0.24\textwidth}
         \centering
         \includegraphics[width=\textwidth]{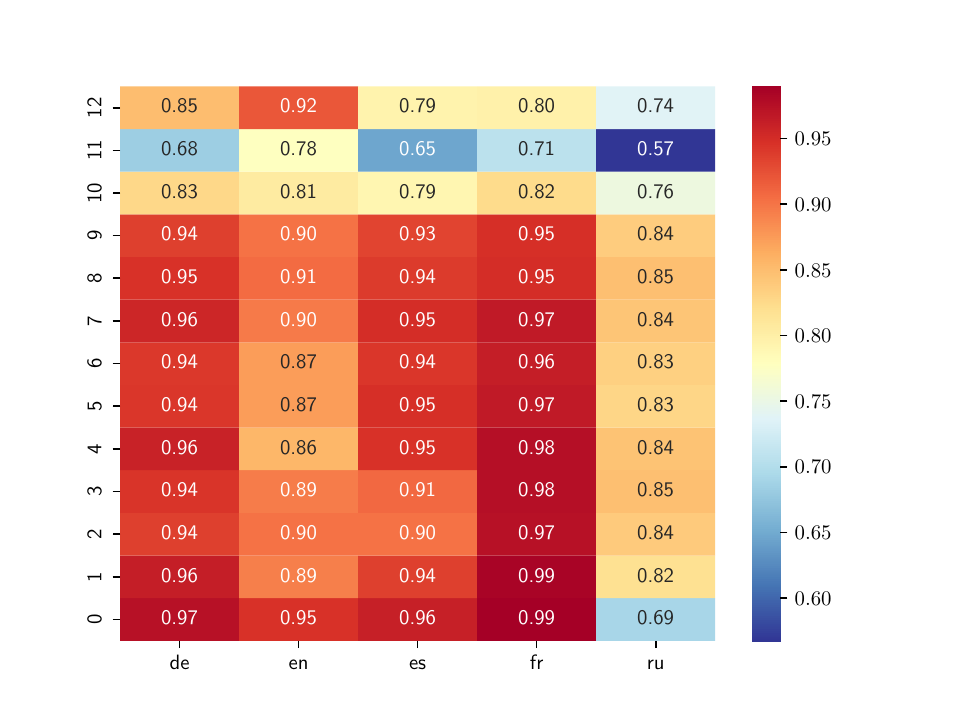}
         \caption{mBERT vs mBART m2m}
         \label{fig:bert-vs-m2m}
     \end{subfigure}
     \hfill
     \begin{subfigure}[b]{0.23\textwidth}
         \centering
         \includegraphics[width=\textwidth]{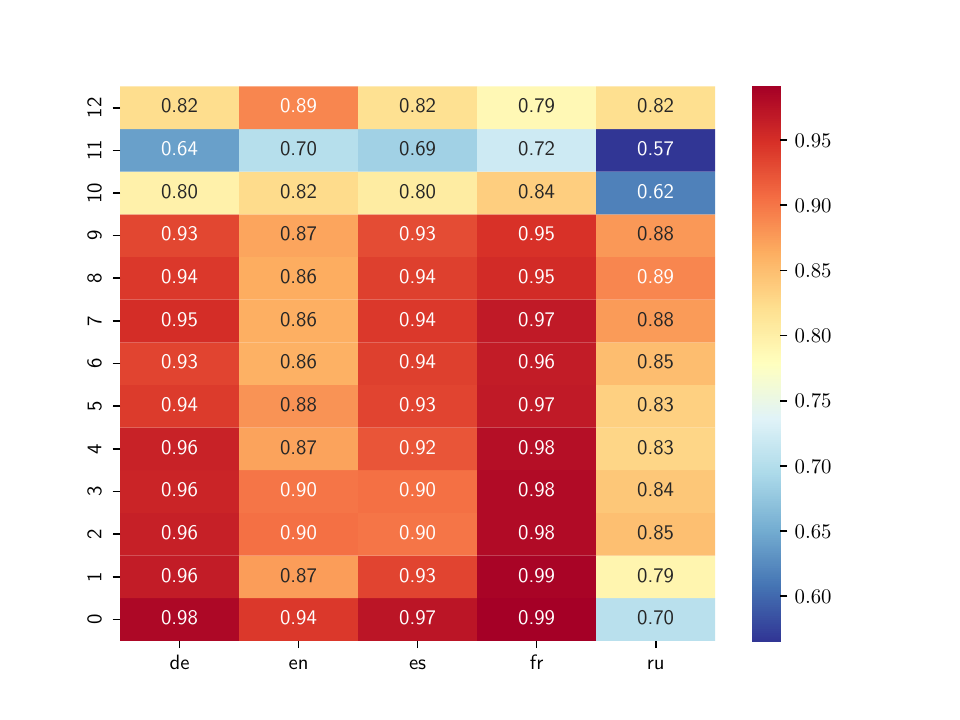}
         \caption{mBERT vs mBART m2o}
         \label{fig:bert-vs-m2o}
     \end{subfigure}
     \hfill
     \begin{subfigure}[b]{0.23\textwidth}
         \centering
         \includegraphics[width=\textwidth]{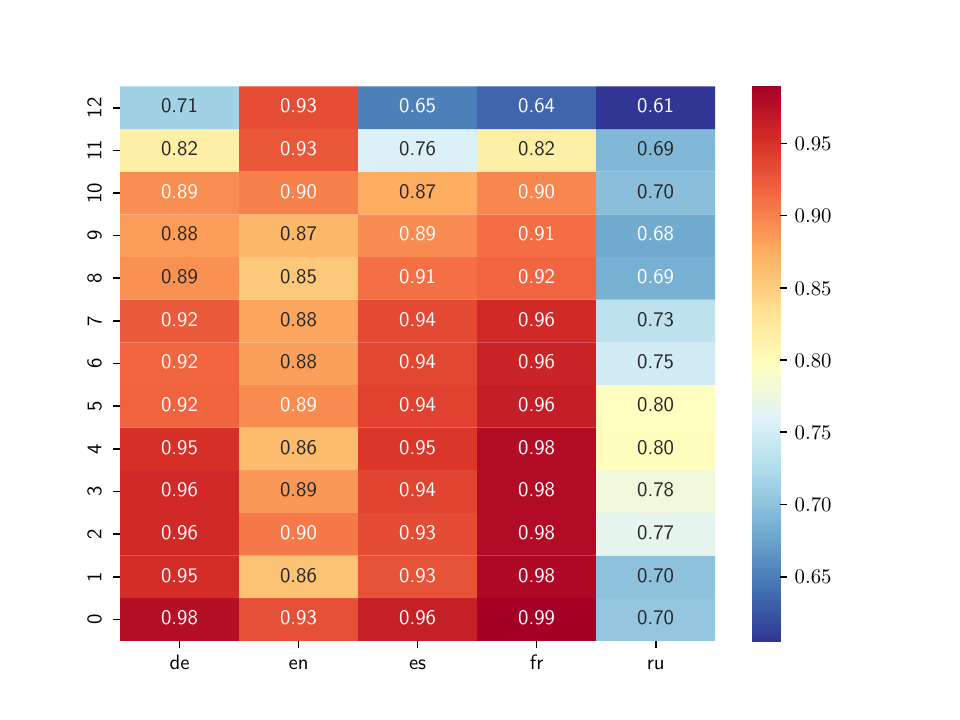}
         \caption{mBERT vs mBART o2m}
         \label{fig:bert-vs-o2m}
     \end{subfigure}
\caption{Representational similarity between mBART-based MT models and LMs}
\label{fig:cka_similarity_between_models}
\end{figure}
\begin{table*}[!hbt]
\centering
\scriptsize
\begin{tabular}{l |cc | cc | cc| cc| cc| cc } 
\toprule
\multirow{2}{*}{Model} & \multicolumn{2}{c|}{K} & \multicolumn{2}{c|}{Q} & \multicolumn{2}{c|}{V} & \multicolumn{2}{c|}{Out} & \multicolumn{2}{c|}{FC up} & \multicolumn{2}{c}{FC down} \\ 
& $\|\sigma\|$ & $d$ & $\|\sigma\|$ & $d$ & $\|\sigma\|$ & $d$ & $\|\sigma\|$ & $d$ & $\|\sigma\|$ & $d$ & $\|\sigma\|$ & $d$ \\ 
\midrule 
mBART  &  44.76  &  --  &  44.85  &  --  &  53.73  &  --  &  53.45  &  --  &  90.25  &  --  &  99.63  &  -- \\ 
\midrule
mBART m2m  &  48.28  &  4.23  &  48.29  &  4.07  &  55.65  &  2.73  &  55.14  &  3.01  &  99.28  &  9.47  &  107.94  &  9.63 \\ 
mBART m2o  &  48.34  &  4.23  &  48.35  &  4.06  &  56.19  &  2.95  &  55.73  &  2.99  &  101.06  &  11.19  &  109.71  &  11.18 \\ 
mBART o2m  &  56.13  &  11.76  &  56.25  &  11.74  &  60.17  &  7.18  &  59.32  &  7.07  &  116.17  &  26.34  &  120.50  &  22.15 \\ 
\bottomrule 
\end{tabular}
\caption{SVD scaling effect for mBART and  CP models; weight matrices from the 12th layer.}
\label{tab:weight_layer12}
\end{table*}
We first examine the hidden representations by comparing the representational similarity between different models using the Centered Kernel Alignment (CKA)~\citep{kornblith2019similarity} metric. 
CKA is calculated as 
\begin{equation*}
    \mathrm{CKA}(\mathbf{X}, \mathbf{Y})=\frac{\left\|\mathbf{Y}^{\top} \mathbf{X}\right\|_{\mathrm{F}}^{2}}{\left(\left\|\mathbf{X}^{\top} \mathbf{X}\right\|_{\mathrm{F}}\left\|\mathbf{Y}^{\top} \mathbf{Y}\right\|_{\mathrm{F}}\right)}
\end{equation*}
where $\mathbf{X}\in \mathbb{R}^{n\times d}$ and $\mathbf{Y}\in \mathbb{R}^{n\times d}$ are pooled representations of $n$ data samples with the dimension of $d$, and $\left\|~\cdot~\right\|_\mathrm{F}$ denotes the Frobenius norm.
\Cref{fig:cka_similarity_between_models} shows the representational similarity between CP models (mBART-based multilingual MT) and language models (mBERT, mBART, and XLM-R) obtained from 80 data samples from the NC dataset. 
CP models based on mBART (m2m and m2o) learn more similar language representations to mBART than XLM-R because the MT pretraining of these models was continued from an mBART checkpoint. 
However, some representations of mBART o2m, especially those in Russian, are highly dissimilar to those of mBART. 
We assume this is an effect of the continued pretraining with a translation objective from English to other languages: Cyrillic script being irrelevant to this task, we can expect that the o2m CP model does not need to maintain the quality of the corresponding word-piece representations.
We also observe some outliers in the representational similarity. 
Two CP models (m2m and m2o) also learn more distinct representations for German in the 10th and 11th layers.

\begin{figure*}[h]
     \centering
     \begin{subfigure}[b]{\textwidth}
         \centering
         \includegraphics[width=\textwidth]{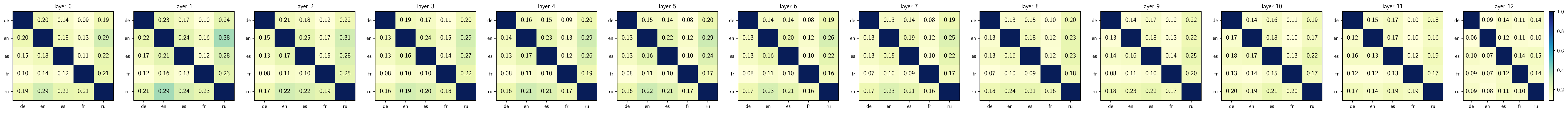}
         \caption{mBART (lower triangle) and mBART m2m (upper triangle)}
         \label{fig:mbart-m2m}
     \end{subfigure}
     \hfill
     \begin{subfigure}[b]{\textwidth}
         \centering
         \includegraphics[width=\textwidth]{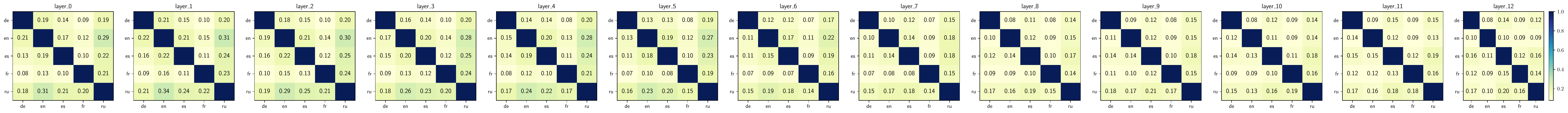}
         \caption{mBART m2o (lower triangle) and mBART o2m (upper triangle)}
         \label{fig:m2o-vs-o2m}
     \end{subfigure}
     \hfill
     \begin{subfigure}[b]{\textwidth}
         \centering
         \includegraphics[width=\textwidth]{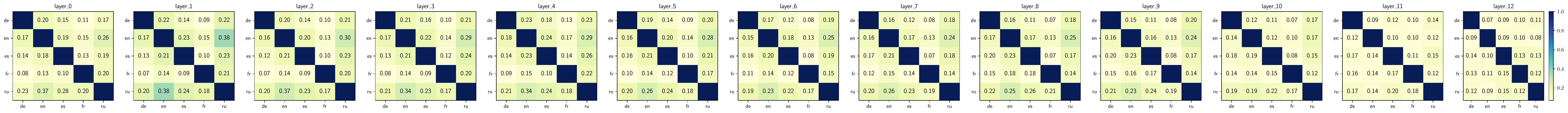}
         \caption{mBERT (lower triangle) and XLM-R (upper triangle)}
         \label{fig:bert-xlmr}
     \end{subfigure}
\caption{Representational similarity between different languages with representations learned by LMs and MT models}
\label{fig:cka_similarity_between_languages}
\end{figure*}

We now turn to the representational similarity between language pairs. 
\Cref{fig:cka_similarity_between_languages} shows the representational similarity between language pairs learned by MT models and LMs. 
The results suggest that LMs learn more language-agnostic representations than MT models

In all, we cannot establish a strong connection between representational similarity and downstream performance. 
Instead, we see a trend that CP models maintain comparatively similar representations to the model they are derived from and dissimilar representations to other LMs.

\subsection{CP's effect on scaling}
\label{sec:scaling}
We have established in the previous \Cref{sec:RSA} that variations in the representation space are mostly tied to the sequence of updates performed on some architecture.
Explaining why MT objectives fail to enhance performances on cross-lingual tasks, therefore, requires that we study the actual computations being performed by CP models.
Hence, we investigate the weight matrices of the mBART LM and CP models.
Our intuition is as follows: 
Weight matrices are linear maps, and we can make some sense of the specific characteristics of these maps.
More precisely, we focus on the magnitude of the eigenvalues: Higher absolute values of eigenvalues should entail numerically larger component values in the output vectors. 
Intuitively, this should impact how separable the output vectors are after applying the weight matrix transformation, which should be encoded in the corresponding eigenvalues.
\footnote{
    This last expectation of separability might not be borne out if the inputs are proportionally less separable in models with larger eigenvalues. In Transformers, this ultimately depends on the eigenvalues of the embedding weights and the (non-linear) computations performed in earlier layers. We leave this aspect for future work.
}
In practice, we apply singular value decomposition (SVD) to retrieve singular values instead:
weight matrices $\mathbf{W}$ can be rewritten as $\mathbf{W}=\Sigma_{i=1}^{r}\sigma_i\mathbf{u}_i\mathbf{v}_{i}^\top$, where $\mathbf{u}_i$ and $\mathbf{v}_{i}$ form two sets of orthogonal vectors, which are combined through the scaling factors $\sigma_i$, known as singular values.

We analyze the scaling effect with or without MT as continued pretraining by comparing mBART-based multilingual machine translation models with the mBART language model. 
We compute the singular values of the weight matrices for key, query, value, and output projections in the transformer multi-head attention sub-layers and both up and down projection weight matrices of the fully connected (FC) layers.
After decomposition, we calculate the norm of the vector $\mathbf{\sigma} = (\sigma_1 , \dots, \sigma_r)$ of singular values, which we denote as $\|\sigma\|$, as well as the difference of singular values in mBART and the CP models (denoted as $d$). 
\Cref{tab:weight_layer12} reports the values of the 12th layer.\footnote{Results for all other layers are available in \Cref{sec:scaling_additional}, \Cref{tab:scaling_all_layers}.}

CP models have larger singular values than the mBART they are derived from, therefore having a stronger scaling effect geometrically. 
Also, remark that translation direction in CP impacts singular value differences: o2m lays noticeably further away from mBART than m2m and m2o. 
In all, this suggests that models trained on the MT objective learn to spread their outputs on larger output vector spaces.
We hypothesize that this behavior is helpful for MT as it entails that outputs are more easily separable (as noted by, e.g., \citealp{chen2023offtarget}); but it might also hinder downstream performances by making the model harder to adapt to other tasks where such behavior is unnecessary or detrimental.

\section{Related works} %
Pretrained language models such as BERT show surprising performance in cross-lingual tasks~\citep{wu2019beto}, a domain that is intensively studied and exhibits various applications~\citep{pikuliak2021cross}.
\citet{huang-etal-2019-unicoder} further enhanced LM cross-lingual performances via universal language encoding.
\citet{eriguchi2018zeroshot} conducted an early study on using the encoder of multilingual MT models for three cross-lingual classification tasks in high-resource languages. %
Similarly, \citet{chen-etal-2021-zero} utilized pretrained multilingual MT encoders and the embedding layers of XLM-R to propose a two-stage training scheme, yielding improved performance on zero-shot cross-lingual machine translation.
\citet{kale-etal-2021-nmt5} investigate using parallel data for pretraining language models to solve multilingual NLP tasks. 
Our study differs from this work in the following three aspects. 
First, objective and hypothesis. 
While both studies involve incorporating parallel data into pre-training, the starting hypotheses and objectives differ. \citet{kale-etal-2021-nmt5} explored the general benefits of pre-training with parallel data, whereas we specifically investigate the impact of an MT objective on cross-lingual transfer.
Second, methodology. Our study introduces the concept of continued pre-training (CP) as a sequential training process specifically focused on the MT objective. In contrast, \citet{kale-etal-2021-nmt5} performed multi-tasking during pre-training with various objectives, including machine translation.
Third, model configurations. We use mBART models with different translation settings for CP, while \citet{kale-etal-2021-nmt5} focused on mT5 as a massively multilingual model.
More broadly, previous studies have leveraged pretrained encoder-decoder LMs to build effective MT models \citep{liu-etal-2020-multilingual-denoising,tang2020multilingual}, which suggests that MT and LM are not entirely unrelated tasks---though the evidence is conflicting \citep{vazquez-etal-2021-differences}.

\section{Conclusion}
\label{sec:conclusion}

This paper reports empirical studies on cross-lingual transfer learning using existing pretrained multilingual language and machine translation models. 
We have investigated whether machine translation as continued pretraining can bridge multilingual pretraining and cross-lingual transfer learning. 
Our empirical results in \Cref{sec:perfs} showed that CP with the MT objective failed to improve cross-lingual performance. %
Further analyses of the language representations learned by different models in \Cref{sec:RSA} and of their weight matrices %
in \Cref{sec:scaling}
showed that models re-trained on the MT objective display larger scaling factors than the checkpoint they were derived from, suggesting that machine translation fosters output separability.
Simply put, models trained on MT objectives need not have representations that match those of multilingual LMs that succeed on cross-lingual transfer tasks: What is useful for MT may be detrimental in other cross-lingual downstream applications.
Our objective was to shed light on potential pitfalls or challenges associated with additional translation task learning in multilingual language models, rather than strictly aiming for performance improvement. The identified relationship between changes in distributional representations and performance degradation is a valuable insight that contributes to our understanding of model behavior in multilingual scenarios.

In future work, we intend to pursue two distinct directions: (i) establishing a principled comparison instead of relying on publicly available pretrained models to more accurately control for parameter count, architecture design, and training data; (ii) studying more formally to what extent separability in MT is attested and distinct from what we observe in LMs.

\section*{Ethical Consideration and Limitations}
We believe this work to comply with all ethical standards.

The present study was not conducted in rigorously comparable settings, such as ensuring that models are exposed to the same pretraining data. 
This limits our capacity to ensure fair comparisons:
\begin{itemize}
    \item In the empirical comparison between language models and continued-trained machine translation models, the training corpora of those models vary. Especially, mBART has not seen some languages in the downstream benchmarks. 
    \item mBART series models have 10\% parameters than BERT and XLM-R, making the comparison in \Cref{tab:mt-continued-cross-lingual} unfair. Nevertheless, mBART encoders did not benefit from the increased number of parameters and failed to achieve better performance in cross-lingual tasks.
\end{itemize}

\section*{Acknowledgments}

This work is part of the FoTran project, funded by the European Research Council (ERC) under the EU's Horizon 2020 research and innovation program (agreement \textnumero{}~771113) and has received funding from the European Union’s Horizon Europe research and innovation programme under Grant agreement No 101070350 and from UK Research and Innovation (UKRI) under the UK government’s Horizon Europe funding guarantee [grant number 10052546]. 
The authors wish to acknowledge CSC – IT Center for Science, Finland, for generous computational resources, including Puhti, Mahti, and LUMI extreme scale access (MOOMIN and LumiNMT).

\section{Bibliographical References}\label{sec:reference}

\bibliographystyle{lrec-coling2024-natbib}
\bibliography{lm-vs-mt}

\begin{thebibliography}{28}
\expandafter\ifx\csname natexlab\endcsname\relax\def\natexlab#1{#1}\fi

\bibitem[{Cao et~al.(2020)Cao, Kitaev, and Klein}]{cao2020multilingual}
Steven Cao, Nikita Kitaev, and Dan Klein. 2020.
\newblock Multilingual alignment of contextual word representations.
\newblock In \emph{International Conference on Learning Representations}.

\bibitem[{Chen et~al.(2021)Chen, Ma, Chen, Dong, Zhang, Pan, Wang, and
  Wei}]{chen-etal-2021-zero}
Guanhua Chen, Shuming Ma, Yun Chen, Li~Dong, Dongdong Zhang, Jia Pan, Wenping
  Wang, and Furu Wei. 2021.
\newblock \href {https://doi.org/10.18653/v1/2021.emnlp-main.2} {Zero-shot
  cross-lingual transfer of neural machine translation with multilingual
  pretrained encoders}.
\newblock In \emph{Proceedings of the 2021 Conference on Empirical Methods in
  Natural Language Processing}, pages 15--26, Online and Punta Cana, Dominican
  Republic. Association for Computational Linguistics.

\bibitem[{Chen et~al.(2023)Chen, Ma, Zhang, Wei, and Chang}]{chen2023offtarget}
Liang Chen, Shuming Ma, Dongdong Zhang, Furu Wei, and Baobao Chang. 2023.
\newblock \href {https://doi.org/10.18653/v1/2023.findings-acl.608} {On the
  off-target problem of zero-shot multilingual neural machine translation}.
\newblock In \emph{Findings of the Association for Computational Linguistics:
  ACL 2023}, pages 9542--9558, Toronto, Canada. Association for Computational
  Linguistics.

\bibitem[{Conneau et~al.(2020)Conneau, Khandelwal, Goyal, Chaudhary, Wenzek,
  Guzm{\'a}n, Grave, Ott, Zettlemoyer, and Stoyanov}]{conneau2020unsupervised}
Alexis Conneau, Kartikay Khandelwal, Naman Goyal, Vishrav Chaudhary, Guillaume
  Wenzek, Francisco Guzm{\'a}n, {\'E}douard Grave, Myle Ott, Luke Zettlemoyer,
  and Veselin Stoyanov. 2020.
\newblock Unsupervised cross-lingual representation learning at scale.
\newblock In \emph{Proceedings of the 58th Annual Meeting of the Association
  for Computational Linguistics}, pages 8440--8451.

\bibitem[{Conneau et~al.(2018)Conneau, Lample, Rinott, Williams, Bowman,
  Schwenk, and Stoyanov}]{conneau2018XNLIEC}
Alexis Conneau, Guillaume Lample, Ruty Rinott, Adina Williams, Samuel~R.
  Bowman, Holger Schwenk, and Veselin Stoyanov. 2018.
\newblock {XNLI: Evaluating Cross-lingual Sentence Representations}.
\newblock In \emph{EMNLP}.

\bibitem[{Costa-juss{\`a} et~al.(2022)Costa-juss{\`a}, Cross, {\c{C}}elebi,
  Elbayad, Heafield, Heffernan, Kalbassi, Lam, Licht, Maillard
  et~al.}]{costa2022no}
Marta~R Costa-juss{\`a}, James Cross, Onur {\c{C}}elebi, Maha Elbayad, Kenneth
  Heafield, Kevin Heffernan, Elahe Kalbassi, Janice Lam, Daniel Licht, Jean
  Maillard, et~al. 2022.
\newblock No language left behind: Scaling human-centered machine translation.
\newblock \emph{arXiv preprint arXiv:2207.04672}.

\bibitem[{Devlin et~al.(2019)Devlin, Chang, Lee, and
  Toutanova}]{devlin2019bert}
Jacob Devlin, Ming-Wei Chang, Kenton Lee, and Kristina Toutanova. 2019.
\newblock {BERT: Pre-training of Deep Bidirectional Transformers for Language
  Understanding}.
\newblock In \emph{Proceedings of the 2019 Conference of the North American
  Chapter of the Association for Computational Linguistics: Human Language
  Technologies}.

\bibitem[{Eriguchi et~al.(2018)Eriguchi, Johnson, Firat, Kazawa, and
  Macherey}]{eriguchi2018zeroshot}
Akiko Eriguchi, Melvin Johnson, Orhan Firat, Hideto Kazawa, and Wolfgang
  Macherey. 2018.
\newblock \href {http://arxiv.org/abs/1809.04686} {Zero-shot cross-lingual
  classification using multilingual neural machine translation}.

\bibitem[{Huang et~al.(2019)Huang, Liang, Duan, Gong, Shou, Jiang, and
  Zhou}]{huang-etal-2019-unicoder}
Haoyang Huang, Yaobo Liang, Nan Duan, Ming Gong, Linjun Shou, Daxin Jiang, and
  Ming Zhou. 2019.
\newblock \href {https://doi.org/10.18653/v1/D19-1252} {{U}nicoder: A universal
  language encoder by pre-training with multiple cross-lingual tasks}.
\newblock In \emph{Proceedings of the 2019 Conference on Empirical Methods in
  Natural Language Processing and the 9th International Joint Conference on
  Natural Language Processing (EMNLP-IJCNLP)}, pages 2485--2494, Hong Kong,
  China. Association for Computational Linguistics.

\bibitem[{Kale et~al.(2021)Kale, Siddhant, Al-Rfou, Xue, Constant, and
  Johnson}]{kale-etal-2021-nmt5}
Mihir Kale, Aditya Siddhant, Rami Al-Rfou, Linting Xue, Noah Constant, and
  Melvin Johnson. 2021.
\newblock \href {https://doi.org/10.18653/v1/2021.acl-short.87} {nm{T}5 - is
  parallel data still relevant for pre-training massively multilingual language
  models?}
\newblock In \emph{Proceedings of the 59th Annual Meeting of the Association
  for Computational Linguistics and the 11th International Joint Conference on
  Natural Language Processing (Volume 2: Short Papers)}, pages 683--691,
  Online. Association for Computational Linguistics.

\bibitem[{Kornblith et~al.(2019)Kornblith, Norouzi, Lee, and
  Hinton}]{kornblith2019similarity}
Simon Kornblith, Mohammad Norouzi, Honglak Lee, and Geoffrey Hinton. 2019.
\newblock Similarity of neural network representations revisited.
\newblock In \emph{International Conference on Machine Learning}, pages
  3519--3529. PMLR.

\bibitem[{Liang et~al.(2020)Liang, Duan, Gong, Wu, Guo, Qi, Gong, Shou, Jiang,
  Cao, Fan, Zhang, Agrawal, Cui, Wei, Bharti, Qiao, Chen, Wu, Liu, Yang,
  Campos, Majumder, and Zhou}]{liang2020XGLUE}
Yaobo Liang, Nan Duan, Yeyun Gong, Ning Wu, Fenfei Guo, Weizhen Qi, Ming Gong,
  Linjun Shou, Daxin Jiang, Guihong Cao, Xiaodong Fan, Ruofei Zhang, Rahul
  Agrawal, Edward Cui, Sining Wei, Taroon Bharti, Ying Qiao, Jiun-Hung Chen,
  Winnie Wu, Shuguang Liu, Fan Yang, Daniel Campos, Rangan Majumder, and Ming
  Zhou. 2020.
\newblock {XGLUE}: A new benchmark dataset for cross-lingual pre-training,
  understanding and generation.
\newblock \emph{arXiv}, abs/2004.01401.

\bibitem[{Liu et~al.(2020)Liu, Gu, Goyal, Li, Edunov, Ghazvininejad, Lewis, and
  Zettlemoyer}]{liu-etal-2020-multilingual-denoising}
Yinhan Liu, Jiatao Gu, Naman Goyal, Xian Li, Sergey Edunov, Marjan
  Ghazvininejad, Mike Lewis, and Luke Zettlemoyer. 2020.
\newblock \href {https://doi.org/10.1162/tacl_a_00343} {Multilingual denoising
  pre-training for neural machine translation}.
\newblock \emph{Transactions of the Association for Computational Linguistics},
  8:726--742.

\bibitem[{Loshchilov and Hutter(2019)}]{loshchilov2019decoupled}
Ilya Loshchilov and Frank Hutter. 2019.
\newblock \href {http://arxiv.org/abs/1711.05101} {Decoupled weight decay
  regularization}.

\bibitem[{Lu et~al.(2018)Lu, Keung, Ladhak, Bhardwaj, Zhang, and
  Sun}]{lu-etal-2018-neural}
Yichao Lu, Phillip Keung, Faisal Ladhak, Vikas Bhardwaj, Shaonan Zhang, and
  Jason Sun. 2018.
\newblock \href {https://doi.org/10.18653/v1/W18-6309} {A neural interlingua
  for multilingual machine translation}.
\newblock In \emph{Proceedings of the Third Conference on Machine Translation:
  Research Papers}, pages 84--92, Brussels, Belgium. Association for
  Computational Linguistics.

\bibitem[{Masterman(1961)}]{masterman-1961-semantic}
Margaret Masterman. 1961.
\newblock \href {https://aclanthology.org/1961.earlymt-1.24} {Semantic message
  detection for machine translation, using an interlingua}.
\newblock In \emph{Proceedings of the International Conference on Machine
  Translation and Applied Language Analysis}, National Physical Laboratory,
  Teddington, UK.

\bibitem[{Pikuliak et~al.(2021)Pikuliak, {\v{S}}imko, and
  Bielikov{\'a}}]{pikuliak2021cross}
Mat{\'u}{\v{s}} Pikuliak, Mari{\'a}n {\v{S}}imko, and M{\'a}ria Bielikov{\'a}.
  2021.
\newblock Cross-lingual learning for text processing: A survey.
\newblock \emph{Expert Systems with Applications}, 165:113765.

\bibitem[{Sang(2002)}]{sang2002IntroductionTT}
Erik F. Tjong~Kim Sang. 2002.
\newblock Introduction to the conll-2002 shared task: Language-independent
  named entity recognition.
\newblock \emph{ArXiv}, cs.CL/0209010.

\bibitem[{Sang and Meulder(2003)}]{sang2003IntroductionTT}
Erik F. Tjong~Kim Sang and Fien~De Meulder. 2003.
\newblock Introduction to the conll-2003 shared task: Language-independent
  named entity recognition.
\newblock \emph{ArXiv}, cs.CL/0306050.

\bibitem[{Tang et~al.(2020)Tang, Tran, Li, Chen, Goyal, Chaudhary, Gu, and
  Fan}]{tang2020multilingual}
Yuqing Tang, Chau Tran, Xian Li, Peng-Jen Chen, Naman Goyal, Vishrav Chaudhary,
  Jiatao Gu, and Angela Fan. 2020.
\newblock Multilingual translation with extensible multilingual pretraining and
  finetuning.
\newblock \emph{arXiv preprint arXiv:2008.00401}.

\bibitem[{Tiedemann(2018)}]{Tiedemann2018EmergingLS}
J{\"o}rg Tiedemann. 2018.
\newblock \href
  {https://www.helsinki.fi/en/helsinki-centre-for-digital-humanities/dhn-2018}
  {Emerging language spaces learned from massively multilingual corpora}.
\newblock In \emph{Proceedings of the Digital Humanities in the Nordic
  Countries 3rd Conference (DHN 2018)}, volume 2084 of \emph{CEUR Workshop
  Proceedings}, pages 188--197, Unknown. CEUR Workshop Proceedings.
\newblock Digital humanities in the Nordic Countries DHN2018, DHN2018 ;
  Conference date: 07-03-2018 Through 09-03-2018.

\bibitem[{V{\'a}zquez et~al.(2021)V{\'a}zquez, Celikkanat, Creutz, and
  Tiedemann}]{vazquez-etal-2021-differences}
Ra{\'u}l V{\'a}zquez, Hande Celikkanat, Mathias Creutz, and J{\"o}rg Tiedemann.
  2021.
\newblock \href {https://doi.org/10.18653/v1/2021.acl-srw.35} {On the
  differences between {BERT} and {MT} encoder spaces and how to address them in
  translation tasks}.
\newblock In \emph{Proceedings of the 59th Annual Meeting of the Association
  for Computational Linguistics and the 11th International Joint Conference on
  Natural Language Processing: Student Research Workshop}, pages 337--347,
  Online. Association for Computational Linguistics.

\bibitem[{Wang et~al.(2019)Wang, Che, Guo, Liu, and Liu}]{wang2019cross}
Yuxuan Wang, Wanxiang Che, Jiang Guo, Yijia Liu, and Ting Liu. 2019.
\newblock Cross-lingual bert transformation for zero-shot dependency parsing.
\newblock In \emph{Proceedings of the 2019 Conference on Empirical Methods in
  Natural Language Processing and the 9th International Joint Conference on
  Natural Language Processing (EMNLP-IJCNLP)}, pages 5721--5727.

\bibitem[{Wu and Dredze(2019)}]{wu2019beto}
Shijie Wu and Mark Dredze. 2019.
\newblock Beto, bentz, becas: The surprising cross-lingual effectiveness of
  bert.
\newblock In \emph{Proceedings of the 2019 Conference on Empirical Methods in
  Natural Language Processing and the 9th International Joint Conference on
  Natural Language Processing (EMNLP-IJCNLP)}, pages 833--844.

\bibitem[{Wu and Dredze(2020)}]{wu2020explicit}
Shijie Wu and Mark Dredze. 2020.
\newblock Do explicit alignments robustly improve multilingual encoders?
\newblock In \emph{Proceedings of the 2020 Conference on Empirical Methods in
  Natural Language Processing (EMNLP)}, pages 4471--4482.

\bibitem[{Xue et~al.(2021)Xue, Constant, Roberts, Kale, Al-Rfou, Siddhant,
  Barua, and Raffel}]{xue-etal-2021-mt5}
Linting Xue, Noah Constant, Adam Roberts, Mihir Kale, Rami Al-Rfou, Aditya
  Siddhant, Aditya Barua, and Colin Raffel. 2021.
\newblock \href {https://doi.org/10.18653/v1/2021.naacl-main.41} {m{T}5: A
  massively multilingual pre-trained text-to-text transformer}.
\newblock In \emph{Proceedings of the 2021 Conference of the North American
  Chapter of the Association for Computational Linguistics: Human Language
  Technologies}, pages 483--498, Online. Association for Computational
  Linguistics.

\bibitem[{Yang et~al.(2019)Yang, Zhang, Tar, and Baldridge}]{yang2019PAWSXAC}
Yinfei Yang, Yuan Zhang, Chris Tar, and Jason Baldridge. 2019.
\newblock Paws-x: A cross-lingual adversarial dataset for paraphrase
  identification.
\newblock \emph{ArXiv}, abs/1908.11828.

\bibitem[{Zeman et~al.(2020)Zeman, Nivre, Abrams, Ackermann, Aepli, Aghaei,
  Ziane, and et~al.}]{universal-dependencies}
Daniel Zeman, Joakim Nivre, Mitchell Abrams, Elia Ackermann, No{\"e}mi Aepli,
  Hamid Aghaei, R~Ziane, and et~al. 2020.
\newblock Universal dependencies 2.5.

\end{thebibliography}

\clearpage
\appendix

\section{Benchmarks for downstream evaluation}
\label{sec:detailed_benchmarks}
A summary of benchmarks for downstream evaluation is shown in \Cref{tab:statistics}.

\begin{table*}[ht!]
\small
	\centering
	\begin{tabular}{lcccccccc}
		\toprule
		Task & \# of Languages & $\vert$Train$\vert^{en}$ & $\vert$Dev$\vert^{avg}$ & $\vert$Test$\vert^{avg}$ & Metric & Data Source\\
		\midrule
		\multicolumn{1}{l}{NER}         & 4 &  15.0K & 2.8K & 3.4K & F1 & ECI Multilingual Text Corpus \\
        \multicolumn{1}{l}{POS}         & 18 & 25.4K & 1.0K & 0.9K & ACC & UD Tree-banks~(v2.5)  \\
        \multicolumn{1}{l}{NC}      & 5 & 100K & 10K & 10K & ACC & Commercial News Website \\
        \multicolumn{1}{l}{XNLI}        & 15 & 433K & 2.5K & 5K & ACC & MultiNLI Corpus \\
        \multicolumn{1}{l}{PAWS-X}      & 4 & 49.4K & 2K & 2K & ACC & Wikipedia \\
        \multicolumn{1}{l}{QADSM}   & 3 & 100K & 10K & 10K & ACC & Commercial Search Engine \\
        \multicolumn{1}{l}{WPR}     & 7 & 100K & 10K & 10K & nDCG & Commercial Search Engine \\
        \multicolumn{1}{l}{QAM}     & 3 & 100K & 10K & 10K & ACC & Commercial Search Engine \\
		\bottomrule
	\end{tabular}
	\caption{A summary of benchmarks for downstream evaluation. We choose 8 downstream tasks from XGLUE~\citep{liang2020XGLUE} for cross-lingual evaluation and x tasks for monolingual evaluation. %
	The training set of each task is only available in English, with $\vert$Train$\vert^{en}$ denoting the number of labeled instances. $\vert$Dev$\vert^{avg}$ and $\vert$Test$\vert^{avg}$ denote
	the average numbers of labeled instances in the dev sets and test sets, respectively. }
	\label{tab:statistics}
\end{table*}

\section{Supplementary results}
\subsection{Per-language results of cross-lingual evaluation}
\label{sec:per-language-results}

\Cref{tab:mt-continued-cross-lingual-per-lang} shows the overall performance of cross-lingual evaluation using LMs and MT models. 
Note that models are fine-tuned only in English and evaluated in other languages; moreover benchmarks differ in which languages they include. 
As a consequence, some scores are not available for some languages.

\begin{table*}
\small
\centering
\resizebox{1\linewidth}{!}{
    \begin{tabular}{llcccccccccccccccccccc}
		\toprule
Task	&	Model	&	AR	&	BG	&	DE	&	EL	&	EN	&	ES	&	FR	&	HI	&	IT	&	NL	&	PL	&	PT	&	RU	&	SW	&	TH	&	TR	&	UR	&	VI	&	ZH	&	AVG	\\
\midrule																																											
\multirow{7}{*}{NC}	&	mBERT	&	-	&	-	&	81.09	&	-	&	91.98	&	80.73	&	75.84	&	-	&	-	&	-	&	-	&	-	&	76.96	&	-	&	-	&	-	&	-	&	-	&	-	&	81.32	\\
	&	XLM-R	&	-	&	-	&	81.79	&	-	&	91.97	&	82.14	&	76.10	&	-	&	-	&	-	&	-	&	-	&	78.63	&	-	&	-	&	-	&	-	&	-	&	-	&	82.13	\\
	&	mBART	&	-	&	-	&	82.82	&	-	&	91.90	&	81.37	&	75.95	&	-	&	-	&	-	&	-	&	-	&	78.43	&	-	&	-	&	-	&	-	&	-	&	-	&	82.09	\\
	&	mBART m2o	&	-	&	-	&	80.32	&	-	&	91.46	&	78.95	&	74.11	&	-	&	-	&	-	&	-	&	-	&	76.96	&	-	&	-	&	-	&	-	&	-	&	-	&	80.36	\\
	&	mBART o2m	&	-	&	-	&	63.63	&	-	&	91.39	&	69.12	&	65.40	&	-	&	-	&	-	&	-	&	-	&	37.60	&	-	&	-	&	-	&	-	&	-	&	-	&	65.43	\\
	&	mBART m2m	&	-	&	-	&	77.44	&	-	&	91.78	&	75.02	&	71.99	&	-	&	-	&	-	&	-	&	-	&	75.19	&	-	&	-	&	-	&	-	&	-	&	-	&	78.28	\\
	&	NLLB 600M	&	-	&	-	&	67.52	&	-	&	91.79	&	76.43	&	71.79	&	-	&	-	&	-	&	-	&	-	&	72.41	&	-	&	-	&	-	&	-	&	-	&	-	&	75.99	\\
\hline																																											
\multirow{7}{*}{XNLI}	&	mBERT	&	62.49	&	66.79	&	71.04	&	65.34	&	82.29	&	74.06	&	73.94	&	58.92	&	-	&	-	&	-	&	-	&	65.62	&	51.24	&	51.16	&	61.41	&	56.27	&	68.31	&	68.63	&	65.17	\\
	&	XLM-R	&	70.68	&	76.27	&	76.06	&	74.70	&	84.86	&	79.60	&	78.31	&	68.15	&	-	&	-	&	-	&	-	&	74.18	&	63.78	&	71.29	&	71.89	&	65.34	&	74.10	&	73.21	&	73.49	\\
	&	mBART	&	69.40	&	56.95	&	75.14	&	34.82	&	84.22	&	78.59	&	75.82	&	66.14	&	-	&	-	&	-	&	-	&	73.86	&	58.03	&	67.31	&	68.80	&	62.25	&	71.65	&	71.49	&	67.63	\\
	&	mBART m2o	&	68.96	&	58.76	&	74.90	&	36.87	&	82.13	&	76.59	&	75.62	&	65.78	&	-	&	-	&	-	&	-	&	73.09	&	42.33	&	59.72	&	68.39	&	60.80	&	71.16	&	73.01	&	65.87	\\
	&	mBART o2m	&	42.01	&	43.49	&	49.28	&	34.86	&	81.85	&	53.78	&	55.54	&	40.00	&	-	&	-	&	-	&	-	&	54.70	&	39.80	&	39.84	&	41.57	&	35.86	&	56.02	&	52.45	&	48.07	\\
	&	mBART m2m	&	58.39	&	50.72	&	68.71	&	35.34	&	83.90	&	66.31	&	73.78	&	65.02	&	-	&	-	&	-	&	-	&	57.51	&	40.84	&	49.52	&	60.32	&	58.71	&	70.88	&	62.53	&	60.17	\\
	&	NLLB 600M	&	68.47	&	69.88	&	63.45	&	72.53	&	81.12	&	72.29	&	72.73	&	68.76	&	-	&	-	&	-	&	-	&	72.37	&	58.35	&	64.94	&	59.28	&	64.30	&	68.76	&	67.63	&	68.32	\\
\hline																																											
\multirow{7}{*}{PAWS-X}	&	mBERT	&	-	&	-	&	82.20	&	-	&	92.85	&	84.60	&	86.70	&	-	&	-	&	-	&	-	&	-	&	-	&	-	&	-	&	-	&	-	&	-	&	-	&	86.59	\\
	&	XLM-R	&	-	&	-	&	85.75	&	-	&	93.40	&	88.30	&	87.95	&	-	&	-	&	-	&	-	&	-	&	-	&	-	&	-	&	-	&	-	&	-	&	-	&	88.85	\\
	&	mBART	&	-	&	-	&	86.70	&	-	&	93.65	&	88.30	&	88.30	&	-	&	-	&	-	&	-	&	-	&	-	&	-	&	-	&	-	&	-	&	-	&	-	&	89.24	\\
	&	mBART m2o	&	-	&	-	&	81.80	&	-	&	91.00	&	84.50	&	85.20	&	-	&	-	&	-	&	-	&	-	&	-	&	-	&	-	&	-	&	-	&	-	&	-	&	85.63	\\
	&	mBART o2m	&	-	&	-	&	76.35	&	-	&	89.90	&	78.90	&	81.45	&	-	&	-	&	-	&	-	&	-	&	-	&	-	&	-	&	-	&	-	&	-	&	-	&	81.65	\\
	&	mBART m2m	&	-	&	-	&	83.95	&	-	&	92.20	&	85.75	&	86.80	&	-	&	-	&	-	&	-	&	-	&	-	&	-	&	-	&	-	&	-	&	-	&	-	&	87.18	\\
	&	NLLB 600M	&	-	&	-	&	67.40	&	-	&	82.35	&	71.65	&	72.00	&	-	&	-	&	-	&	-	&	-	&	-	&	-	&	-	&	-	&	-	&	-	&	-	&	73.35	\\
\hline																																											
\multirow{7}{*}{QAM}	&	mBERT	&	-	&	-	&	62.08	&	-	&	69.47	&	-	&	62.35	&	-	&	-	&	-	&	-	&	-	&	-	&	-	&	-	&	-	&	-	&	-	&	-	&	64.63	\\
	&	XLM-R	&	-	&	-	&	66.98	&	-	&	69.69	&	-	&	65.45	&	-	&	-	&	-	&	-	&	-	&	-	&	-	&	-	&	-	&	-	&	-	&	-	&	67.37	\\
	&	mBART	&	-	&	-	&	66.36	&	-	&	70.46	&	-	&	66.71	&	-	&	-	&	-	&	-	&	-	&	-	&	-	&	-	&	-	&	-	&	-	&	-	&	67.84	\\
	&	mBART m2o	&	-	&	-	&	64.20	&	-	&	65.10	&	-	&	62.39	&	-	&	-	&	-	&	-	&	-	&	-	&	-	&	-	&	-	&	-	&	-	&	-	&	63.90	\\
	&	mBART o2m	&	-	&	-	&	55.27	&	-	&	65.41	&	-	&	54.60	&	-	&	-	&	-	&	-	&	-	&	-	&	-	&	-	&	-	&	-	&	-	&	-	&	58.43	\\
	&	mBART m2m	&	-	&	-	&	62.62	&	-	&	66.21	&	-	&	60.72	&	-	&	-	&	-	&	-	&	-	&	-	&	-	&	-	&	-	&	-	&	-	&	-	&	63.18	\\
	&	NLLB 600M	&	-	&	-	&	57.90	&	-	&	66.47	&	-	&	60.14	&	-	&	-	&	-	&	-	&	-	&	-	&	-	&	-	&	-	&	-	&	-	&	-	&	61.50	\\
\hline																																											
\multirow{7}{*}{QADSM}	&	mBERT	&	-	&	-	&	59.94	&	-	&	67.04	&	-	&	62.30	&	-	&	-	&	-	&	-	&	-	&	-	&	-	&	-	&	-	&	-	&	-	&	-	&	63.09	\\
	&	XLM-R	&	-	&	-	&	63.19	&	-	&	71.44	&	-	&	66.02	&	-	&	-	&	-	&	-	&	-	&	-	&	-	&	-	&	-	&	-	&	-	&	-	&	66.88	\\
	&	mBART	&	-	&	-	&	61.83	&	-	&	69.83	&	-	&	64.79	&	-	&	-	&	-	&	-	&	-	&	-	&	-	&	-	&	-	&	-	&	-	&	-	&	65.48	\\
	&	mBART m2o	&	-	&	-	&	63.15	&	-	&	64.07	&	-	&	64.34	&	-	&	-	&	-	&	-	&	-	&	-	&	-	&	-	&	-	&	-	&	-	&	-	&	63.85	\\
	&	mBART o2m	&	-	&	-	&	63.40	&	-	&	65.17	&	-	&	59.61	&	-	&	-	&	-	&	-	&	-	&	-	&	-	&	-	&	-	&	-	&	-	&	-	&	62.73	\\
	&	mBART m2m	&	-	&	-	&	60.89	&	-	&	65.45	&	-	&	62.05	&	-	&	-	&	-	&	-	&	-	&	-	&	-	&	-	&	-	&	-	&	-	&	-	&	62.80	\\
	&	NLLB 600M	&	-	&	-	&	64.48	&	-	&	64.45	&	-	&	62.71	&	-	&	-	&	-	&	-	&	-	&	-	&	-	&	-	&	-	&	-	&	-	&	-	&	63.88	\\
\hline																																											
\multirow{7}{*}{WPR}	&	mBERT	&	-	&	-	&	76.64	&	-	&	77.29	&	75.07	&	73.92	&	-	&	66.58	&	-	&	-	&	77.04	&	-	&	-	&	-	&	-	&	-	&	-	&	62.67	&	74.42	\\
	&	XLM-R	&	-	&	-	&	77.08	&	-	&	77.79	&	76.14	&	74.94	&	-	&	67.87	&	-	&	-	&	77.93	&	-	&	-	&	-	&	-	&	-	&	-	&	62.81	&	75.29	\\
	&	mBART	&	-	&	-	&	76.74	&	-	&	77.18	&	75.41	&	74.22	&	-	&	67.40	&	-	&	-	&	77.38	&	-	&	-	&	-	&	-	&	-	&	-	&	62.86	&	74.72	\\
	&	mBART m2o	&	-	&	-	&	75.60	&	-	&	76.17	&	74.08	&	73.31	&	-	&	66.21	&	-	&	-	&	76.59	&	-	&	-	&	-	&	-	&	-	&	-	&	62.38	&	73.66	\\
	&	mBART o2m	&	-	&	-	&	75.32	&	-	&	75.99	&	74.07	&	72.76	&	-	&	65.39	&	-	&	-	&	75.80	&	-	&	-	&	-	&	-	&	-	&	-	&	61.24	&	73.22	\\
	&	mBART m2m	&	-	&	-	&	76.22	&	-	&	76.22	&	74.28	&	73.23	&	-	&	66.35	&	-	&	-	&	75.79	&	-	&	-	&	-	&	-	&	-	&	-	&	61.93	&	73.68	\\
	&	NLLB 600M	&	-	&	-	&	76.01	&	-	&	76.35	&	73.81	&	73.48	&	-	&	65.84	&	-	&	-	&	76.46	&	-	&	-	&	-	&	-	&	-	&	-	&	62.02	&	73.66	\\
\hline																																											
\multirow{7}{*}{NER}	&	mBERT	&	-	&	-	&	68.84	&	-	&	90.78	&	73.27	&	-	&	-	&	-	&	77.28	&	-	&	-	&	-	&	-	&	-	&	-	&	-	&	-	&	-	&	77.54	\\
	&	XLM-R	&	-	&	-	&	69.99	&	-	&	90.45	&	75.77	&	-	&	-	&	-	&	78.62	&	-	&	-	&	-	&	-	&	-	&	-	&	-	&	-	&	-	&	78.71	\\
	&	mBART	&	-	&	-	&	71.31	&	-	&	91.35	&	72.55	&	-	&	-	&	-	&	75.57	&	-	&	-	&	-	&	-	&	-	&	-	&	-	&	-	&	-	&	77.70	\\
	&	mBART m2o	&	-	&	-	&	52.41	&	-	&	89.61	&	50.71	&	-	&	-	&	-	&	53.36	&	-	&	-	&	-	&	-	&	-	&	-	&	-	&	-	&	-	&	61.52	\\
	&	mBART o2m	&	-	&	-	&	25.66	&	-	&	89.22	&	53.31	&	-	&	-	&	-	&	52.13	&	-	&	-	&	-	&	-	&	-	&	-	&	-	&	-	&	-	&	55.08	\\
	&	mBART m2m	&	-	&	-	&	65.25	&	-	&	88.99	&	66.86	&	-	&	-	&	-	&	66.58	&	-	&	-	&	-	&	-	&	-	&	-	&	-	&	-	&	-	&	71.92	\\<
	&	NLLB 600M	&	-	&	-	&	29.4	&	-	&	89.46	&	43.21	&	-	&	-	&	-	&	54.83	&	-	&	-	&	-	&	-	&	-	&	-	&	-	&	-	&	-	&	54.23	\\
\hline																																											
\multirow{7}{*}{POS}	&	mBERT	&	57.26	&	85.84	&	90.21	&	82.61	&	95.84	&	87.67	&	85.80	&	66.57	&	91.78	&	87.78	&	80.93	&	88.93	&	80.57	&	-	&	41.88	&	68.87	&	60.12	&	55.09	&	60.19	&	76.00	\\
	&	XLM-R	&	69.44	&	88.70	&	91.75	&	87.63	&	96.43	&	88.20	&	89.22	&	72.10	&	91.35	&	88.46	&	83.82	&	90.07	&	87.12	&	-	&	58.08	&	72.76	&	64.28	&	57.06	&	58.45	&	79.72	\\
	&	mBART	&	63.55	&	71.76	&	90.56	&	29.74	&	96.13	&	87.07	&	87.75	&	67.61	&	90.64	&	87.51	&	80.60	&	88.29	&	83.35	&	-	&	55.56	&	66.53	&	55.61	&	54.62	&	51.56	&	72.69	\\
	&	mBART m2o	&	63.97	&	71.30	&	90.64	&	24.82	&	95.74	&	84.98	&	85.19	&	64.32	&	87.45	&	86.18	&	80.12	&	82.91	&	81.91	&	-	&	51.13	&	66.77	&	50.65	&	52.29	&	53.18	&	70.75	\\
	&	mBART o2m	&	53.78	&	58.78	&	61.97	&	41.62	&	95.63	&	63.08	&	70.43	&	48.16	&	59.92	&	60.01	&	53.62	&	58.11	&	61.60	&	-	&	37.12	&	46.21	&	42.90	&	44.95	&	44.08	&	55.67	\\
	&	mBART m2m	&	64.60	&	71.58	&	90.35	&	21.66	&	96.06	&	81.21	&	86.20	&	65.94	&	83.71	&	85.30	&	81.28	&	81.65	&	84.81	&	-	&	41.83	&	63.55	&	52.44	&	51.02	&	51.33	&	69.70	\\
	&	NLLB 600M	&	63.76	&	84.47	&	77.64	&	79.22	&	96.12	&	82.61	&	83.36	&	66.59	&	84.93	&	75.9	&	74.57	&	80.95	&	80.92	&	-	&	46.56	&	56.46	&	58.59	&	45.63	&	46.32	&	71.37	\\
 \bottomrule
    \end{tabular}
}
\caption{The overall performance of cross-lingual natural language understanding. We use the base architecture for mBERT and XLM-R. mBART models only utilize the 12-layer encoders. `m2o' means many-to-one. `o2m' means one-to-many. `m2m' means many-to-many. `-' denotes that the benchmark does not cover the corresponding language.} 
\label{tab:mt-continued-cross-lingual-per-lang} 
\end{table*}

\subsection{The scaling effect of MT as continued training in different layers}
\label{sec:scaling_additional}
\Cref{tab:scaling_all_layers} shows the results of the first 11 layers in the encoders of mBART series models. 
Similar to the analysis in \Cref{sec:scaling}, we calculate the norm of the vectorized diagonal matrices of singular values $\mathrm{diag(\Sigma)}$ and their pairwise distance to the corresponding vectors derived from the same weight matrices in The transformer attention module and fully connected layers in the base mBART model. 
The results indicate the same conclusion drawn from the analysis on the 12th layer. 

\begin{table*}
\centering
\small
\resizebox{1\linewidth}{!}{
\begin{tabular}{l|l |cc | cc | cc| cc| cc| cc }
\toprule 
\multirow{2}{*}{Layer} & \multirow{2}{*}{Model} & \multicolumn{2}{c|}{K} & \multicolumn{2}{c|}{Q} & \multicolumn{2}{c|}{V} & \multicolumn{2}{c|}{Out} & \multicolumn{2}{c|}{FC1} & \multicolumn{2}{c}{FC2} \\ 
& & $\|\sigma\|$ & d & $\|\sigma\|$ & d & $\|\sigma\|$ & d & $\|\sigma\|$ & d  & $\|\sigma\|$ & d & $\|\sigma\|$ & d \\ 
\midrule 
\multirow{4}{*}{0} &  mBART  &  39.25  &  0.00  &  41.16  &  0.00  &  32.82  &  0.00  &  30.30  &  0.00  &  88.19  &  0.00  &  66.57  &  0.00 \\ 
 &  mBART m2m  &  48.46  &  9.55  &  48.83  &  7.71  &  31.01  &  2.14  &  29.29  &  3.10  &  93.43  &  6.64  &  73.70  &  8.30 \\ 
 &  mBART m2o  &  48.49  &  9.59  &  48.95  &  7.90  &  30.51  &  2.44  &  28.99  &  3.55  &  93.94  &  7.12  &  74.39  &  9.03 \\ 
 &  mBART o2m  &  49.15  &  10.62  &  50.54  &  10.50  &  35.90  &  3.87  &  35.75  &  6.73  &  111.30  &  23.86  &  94.52  &  28.87 \\ 
\midrule 
\multirow{4}{*}{1} &  mBART  &  44.41  &  0.00  &  46.89  &  0.00  &  29.21  &  0.00  &  31.80  &  0.00  &  90.87  &  0.00  &  69.86  &  0.00 \\ 
 &  mBART m2m  &  51.21  &  7.58  &  52.62  &  5.79  &  27.69  &  1.73  &  30.56  &  2.41  &  96.44  &  6.34  &  76.99  &  8.09 \\ 
 &  mBART m2o  &  51.64  &  8.07  &  53.10  &  6.36  &  27.49  &  1.78  &  30.47  &  2.63  &  97.06  &  7.31  &  77.83  &  9.07 \\ 
 &  mBART o2m  &  54.00  &  9.77  &  56.38  &  9.70  &  35.17  &  6.40  &  37.90  &  6.64  &  113.14  &  23.36  &  96.06  &  26.69 \\ 
\midrule 
\multirow{4}{*}{2} &  mBART  &  46.56  &  0.00  &  47.80  &  0.00  &  31.80  &  0.00  &  32.22  &  0.00  &  93.76  &  0.00  &  71.51  &  0.00 \\ 
 &  mBART m2m  &  51.83  &  5.86  &  52.66  &  5.04  &  31.34  &  1.32  &  31.96  &  2.48  &  99.62  &  6.49  &  78.89  &  8.19 \\ 
 &  mBART m2o  &  52.40  &  6.53  &  53.22  &  5.66  &  31.26  &  1.09  &  31.94  &  2.53  &  100.24  &  7.50  &  79.71  &  9.18 \\ 
 &  mBART o2m  &  56.22  &  9.75  &  57.47  &  9.91  &  38.20  &  6.91  &  39.20  &  7.45  &  114.89  &  21.77  &  96.37  &  25.03 \\ 
\midrule 
\multirow{4}{*}{3} &  mBART  &  48.09  &  0.00  &  48.65  &  0.00  &  37.77  &  0.00  &  36.85  &  0.00  &  96.27  &  0.00  &  73.45  &  0.00 \\ 
 &  mBART m2m  &  52.57  &  4.54  &  53.05  &  4.63  &  39.06  &  1.65  &  38.06  &  2.42  &  101.80  &  6.09  &  80.15  &  7.27 \\ 
 &  mBART m2o  &  53.15  &  5.15  &  53.62  &  5.21  &  39.15  &  1.59  &  38.12  &  2.40  &  102.31  &  7.18  &  80.84  &  8.12 \\ 
 &  mBART o2m  &  56.82  &  8.80  &  57.69  &  9.28  &  45.90  &  8.25  &  45.43  &  9.00  &  117.17  &  21.21  &  97.57  &  24.24 \\ 
\midrule 
\multirow{4}{*}{4} &  mBART  &  51.59  &  0.00  &  51.06  &  0.00  &  40.80  &  0.00  &  38.91  &  0.00  &  99.22  &  0.00  &  77.23  &  0.00 \\ 
 &  mBART m2m  &  56.45  &  5.05  &  55.97  &  5.08  &  43.02  &  2.51  &  41.20  &  2.71  &  104.97  &  6.76  &  84.00  &  7.36 \\ 
 &  mBART m2o  &  57.10  &  5.70  &  56.62  &  5.72  &  43.00  &  2.31  &  41.17  &  2.50  &  105.74  &  8.23  &  85.01  &  8.72 \\ 
 &  mBART o2m  &  61.88  &  10.40  &  61.58  &  10.61  &  49.68  &  8.98  &  48.54  &  9.82  &  121.06  &  21.94  &  101.56  &  24.42 \\ 
\midrule 
\multirow{4}{*}{5} &  mBART  &  59.93  &  0.00  &  58.27  &  0.00  &  42.09  &  0.00  &  38.29  &  0.00  &  101.79  &  0.00  &  84.25  &  0.00 \\ 
 &  mBART m2m  &  65.19  &  5.44  &  63.76  &  5.69  &  44.03  &  2.11  &  40.23  &  2.61  &  107.87  &  7.42  &  91.05  &  7.56 \\ 
 &  mBART m2o  &  65.86  &  6.14  &  64.44  &  6.33  &  44.22  &  2.21  &  40.42  &  2.63  &  108.94  &  8.86  &  92.49  &  9.56 \\ 
 &  mBART o2m  &  70.59  &  10.94  &  69.24  &  11.26  &  51.39  &  9.36  &  48.64  &  10.70  &  124.78  &  23.18  &  108.30  &  24.15 \\ 
\midrule 
\multirow{4}{*}{6} &  mBART  &  57.55  &  0.00  &  55.80  &  0.00  &  45.27  &  0.00  &  40.54  &  0.00  &  101.52  &  0.00  &  89.89  &  0.00 \\ 
 &  mBART m2m  &  61.71  &  4.60  &  60.09  &  4.47  &  48.11  &  2.94  &  43.59  &  3.34  &  108.45  &  7.71  &  97.85  &  8.11 \\ 
 &  mBART m2o  &  62.22  &  5.12  &  60.62  &  4.93  &  48.44  &  3.24  &  43.95  &  3.63  &  109.45  &  8.79  &  99.23  &  9.60 \\ 
 &  mBART o2m  &  67.76  &  10.73  &  66.31  &  10.74  &  55.16  &  9.96  &  51.59  &  11.24  &  125.45  &  24.20  &  113.90  &  24.10 \\ 
\midrule 
\multirow{4}{*}{7} &  mBART  &  53.94  &  0.00  &  52.44  &  0.00  &  49.99  &  0.00  &  46.20  &  0.00  &  98.28  &  0.00  &  96.88  &  0.00 \\ 
 &  mBART m2m  &  57.72  &  4.33  &  56.30  &  4.04  &  52.77  &  2.98  &  49.09  &  3.36  &  106.68  &  8.96  &  105.70  &  8.94 \\ 
 &  mBART m2o  &  58.03  &  4.61  &  56.64  &  4.31  &  53.23  &  3.32  &  49.58  &  3.72  &  107.94  &  10.21  &  107.26  &  10.47 \\ 
 &  mBART o2m  &  64.09  &  10.79  &  62.87  &  10.64  &  59.23  &  9.49  &  56.08  &  10.17  &  123.84  &  25.99  &  120.47  &  23.76 \\ 
\midrule 
\multirow{4}{*}{8} &  mBART  &  48.50  &  0.00  &  47.79  &  0.00  &  51.22  &  0.00  &  49.37  &  0.00  &  94.92  &  0.00  &  100.91  &  0.00 \\ 
 &  mBART m2m  &  51.54  &  3.69  &  50.86  &  3.51  &  54.22  &  3.17  &  52.37  &  3.14  &  104.18  &  9.91  &  109.97  &  9.35 \\ 
 &  mBART m2o  &  51.67  &  3.75  &  51.00  &  3.60  &  54.76  &  3.63  &  52.96  &  3.67  &  105.67  &  11.31  &  111.65  &  10.98 \\ 
 &  mBART o2m  &  58.53  &  10.57  &  57.90  &  10.39  &  60.62  &  9.65  &  58.81  &  9.60  &  121.89  &  27.64  &  124.33  &  23.66 \\ 
\midrule 
\multirow{4}{*}{9} &  mBART  &  44.39  &  0.00  &  44.78  &  0.00  &  50.78  &  0.00  &  49.83  &  0.00  &  94.01  &  0.00  &  103.16  &  0.00 \\ 
 &  mBART m2m  &  47.47  &  3.73  &  47.82  &  3.55  &  53.04  &  2.78  &  52.13  &  2.95  &  103.53  &  10.31  &  112.32  &  9.64 \\ 
 &  mBART m2o  &  47.57  &  3.76  &  47.92  &  3.56  &  53.55  &  3.13  &  52.61  &  3.15  &  105.10  &  11.78  &  113.97  &  11.19 \\ 
 &  mBART o2m  &  55.12  &  11.12  &  55.55  &  11.12  &  58.83  &  8.49  &  57.86  &  8.42  &  120.99  &  27.80  &  126.63  &  23.92 \\ 
\midrule 
\multirow{4}{*}{10} &  mBART  &  43.50  &  0.00  &  42.87  &  0.00  &  55.41  &  0.00  &  54.97  &  0.00  &  91.92  &  0.00  &  105.62  &  0.00 \\ 
 &  mBART m2m  &  46.93  &  4.66  &  46.28  &  4.31  &  57.64  &  2.89  &  57.30  &  4.13  &  101.92  &  10.67  &  114.82  &  9.94 \\ 
 &  mBART m2o  &  46.88  &  4.50  &  46.24  &  4.23  &  58.24  &  3.25  &  57.93  &  4.21  &  103.56  &  12.19  &  116.47  &  11.45 \\ 
 &  mBART o2m  &  54.68  &  12.00  &  54.22  &  11.91  &  62.95  &  8.24  &  62.28  &  8.64  &  119.35  &  28.09  &  128.21  &  23.29 \\ 
\bottomrule 
\end{tabular}
}
\caption{The scaling effect via singular value decomposition of mBART and the continued-trained MT models.}
\label{tab:scaling_all_layers}
\end{table*}

\end{document}